\documentclass[sigplan,screen]{acmart}
\usepackage{bbm}
\usepackage{ulem}
\usepackage{amsmath}
\DeclareMathOperator*{\argmax}{arg\,max}

\newcommand{\hr}[1]{\textcolor{black}{#1}}

\AtBeginDocument{%
  \providecommand\BibTeX{{%
    \normalfont B\kern-0.5em{\scshape i\kern-0.25em b}\kern-0.8em\TeX}}}

\copyrightyear{2024}
\acmYear{2024}
\setcopyright{rightsretained}
\acmConference[DAC '24]{61st ACM/IEEE Design Automation Conference}{June 23--27, 2024}{San Francisco, CA, USA}
\acmBooktitle{61st ACM/IEEE Design Automation Conference (DAC '24), June 23--27, 2024, San Francisco, CA, USA}
\acmDOI{10.1145/3649329.3658473}
\acmISBN{979-8-4007-0601-1/24/06}

\acmConference[DAC'24]{61st ACM/IEEE Design Automation Conference}{June 23--27, 2024}{San Francisco, CA}

\acmSubmissionID{1122}

\begin{document}


\title{EDGE-LLM: Enabling Efficient Large Language Model Adaptation on Edge Devices via Layerwise Unified Compression and Adaptive Layer Tuning \& Voting}

\author{\small Zhongzhi Yu$^1$, Zheng Wang$^1$, Yuhan Li$^1$, Haoran You$^1$, Ruijie Gao$^1$,  Xiaoya Zhou$^3$, Sreenidhi Reedy Bommu$^1$, Yang (Katie) Zhao$^2$, Yingyan (Celine) Lin$^1$}
\affiliation{%
  \institution{$^1$\textit{Georgia Institute of Technology}, $^2$\textit{University of Minnesota, Twin Cities},
  $^3$\textit{University of California, Santa Barbara}}
  \city{\{zyu401, zwang3478, yli3326, hyou37, eiclab.gatech, sbommu3, celine.lin\}@gatech.edu,\\ yangzhao@umn.edu, xiaoyazhou@umail.ucsb.edu}
  \country{}
}


\renewcommand{\shortauthors}{Zhongzhi Yu, et al.}
\renewcommand{\shorttitle}{Edge-LLM}


\begin{abstract}
Efficient adaption of large language models (LLMs) on edge devices is essential for applications requiring continuous and privacy-preserving adaptation and inference. However, existing tuning techniques fall short because of the high computation and memory overhead\hr{s}. To this end, we introduce a \textbf{computation-} and \textbf{memory-}efficient LLM tuning framework, called Edge-LLM, to facilitate affordable and effective LLM adaptation on edge devices. Specifically, Edge-LLM features three core components: \textbf{(1)} a layer-wise unified compression (LUC) technique to reduce the \textbf{computation} overhead by generating layer-wise pruning sparsity and quantization bit-width policies, \textbf{(2)} an adaptive layer tuning and voting scheme to reduce the \textbf{memory} overhead by reducing the backpropagation depth, and \textbf{(3)} a complementary hardware scheduling strategy to handle the irregular computation patterns introduced by LUC and adaptive layer tuning, 
\hr{thereby achieving efficient computation and data movements.}
\hr{
Extensive experiments demonstrate that Edge-LLM achieves a 2.92$\times$ speed up and a 4$\times$ memory overhead reduction as compared to vanilla tuning methods with a comparable task accuracy.
}
Our code is available at \url{https://github.com/GATECH-EIC/Edge-LLM}
\end{abstract}
\vspace{-5em}




\maketitle

\vspace{-0.8em}
\section{Introduction}
In recent days, large language models (LLMs), such as GPT-4~\cite{bubeck2023sparks}, have shown dominating performance across various applications that revolutionize human life. Following this trend, there is an increasing demand to develop 
\hr{efficient tuning techniques}
for LLMs to enable them on applications that require continuous and privacy-preserving adaptation. However, the massive model size of LLMs hinders directly 
\hr{achieving the LLM adaptation on edge devices}
(e.g., on edge GPUs and smartphones). The challenges are twofold: (1) \textit{the excessive \textbf{computation} overhead} encountered when calculating the forward and backward passes of LLMs~\cite{dettmers2023qlora}, and (2) \textit{the cumbersome \textbf{memory} overhead} 
\hr{introduced for}
storing massive model weights and activations through the tuning process. As shown in recent works~\cite{dettmers2023qlora,liu2023llm}, LLMs are typically tuned on cutting-edge GPUs (e.g., with 40GB or 80GB GPU memory), taking more than a GPU day to complete. Even for the state-of-the-art (SOTA) efficient tuning method, effectively tuning relatively small-scale LLMs (e.g., LLaMA-7B) on edge devices remains impractical~\cite{dettmers2023qlora}.

Although several existing efforts aim to address the aforementioned challenges, each has its own drawbacks. (1) To reduce computation overhead, compressing target LLMs first to reduce the model size is a common approach~\cite{dettmers2023qlora,kim2024memory}. However, \textbf{how to effectively reduce the redundancy of LLMs while maintaining their adaptability is still largely unexplored}~\cite{dettmers2023qlora}.  (2) To mitigate memory overhead, existing methods primarily focus on shortening the backpropagation depth~\cite{zhang2023llama,sung2022lst}. Unfortunately, the reduced backpropagation depth results in only \textbf{a fraction of blocks in LLMs being updated, limiting the achievable performance}.

In this paper, we develop a comprehensive solution to tackle the two aforementioned memory and computation challenges, 
\hr{achieving an effective LLM adaptation.}
Specifically, we make the following contributions.
\begin{itemize}
    \item We propose a comprehensive framework, dubbed Edge-LLM, that tackles the memory and computation challenges of 
    \hr{the LLM adaptation}
    from both algorithm and hardware \hr{perspectives}, enabling \hr{the} effective LLM adaptation on edge devices with limited memory and computation resources. 
    \item \textbf{On the algorithm side}, we accomplish this goal from two directions, each primarily focusing on one of the aforementioned challenges: (1) To reduce the \textbf{computation} overhead, we propose a low-cost layer-wise unified compression (LUC) method based on our empirical observation on LLMs' layer-wise sensitivities to quantization and pruning. (2) To reduce the \textbf{memory} overhead, we introduce an adaptive layer tuning and voting scheme. In adaptive layer tuning, we propose to selectively update distinct segments of the target LLM and reduce the memory footprint by directly connecting the output of the current updating segment to the final layer. Further, in adaptive layer voting, we harness the outputs of different segments of the target LLM by voting for an optimized output.

    \item \textbf{On the hardware side}, to better handle the irregular computation patterns (i.e., diverse layer-wise quantization bit-width, layer-wise pruning sparsity, and LLM segments to update) introduced by the proposed algorithms, we further integrate a complementary hardware scheduling module into Edge-LLM. The hardware scheduling module includes a search space and a search strategy considering potential offloading strategies, computation schedules, and tensor placements, aiming to better convert the theoretical reduction in computation overhead to 
    \hr{the hardware efficiency improvement.}
    \item Experiment results and ablation studies \hr{validate} the effectiveness of our proposed Edge-LLM framework. Specifically, Edge-LLM achieves a 0.70\%$\sim$1.29\% higher MMLU score compared with the baseline methods tuned under the same resource constraints and a comparable perplexity on WikiText-2 as LoRA tuning with a 2.92$\times$ lower latency and a 4$\times$ reduction in memory overhead \hr{during} each iteration. 
\end{itemize}


\vspace{-1em}
\section{Background and Motivation}
\subsection{Efficient Tuning Techniques}
\label{sec:tuninig}
\textbf{Parameter-efficient tuning (PET)} comprises techniques for tuning LLMs to new tasks using a limited number of trainable parameters, typically less than 10\% of the total parameters in the target LLMs~\cite{hu2021lora,sung2022lst,yu2023hint,yu2023master}. It offers two major advantages: (1) reduced storage overhead, facilitating scalable multitask deployment, and (2) a marginal reduction in computation and memory overhead, thanks to the reduced number of trainable parameters~\cite{hu2021lora}. Despite PET's widespread use, directly applying it for on-device LLM adaptation remains impractical due to the remaining memory overhead is still significant. This is because PET typically inserts a learnable adapter to most, if not all, layers of the target LLM, leading to significant memory overhead to store intermediate activations during tuning.

\begin{figure}
    \centering
    \includegraphics[width=0.9\linewidth]{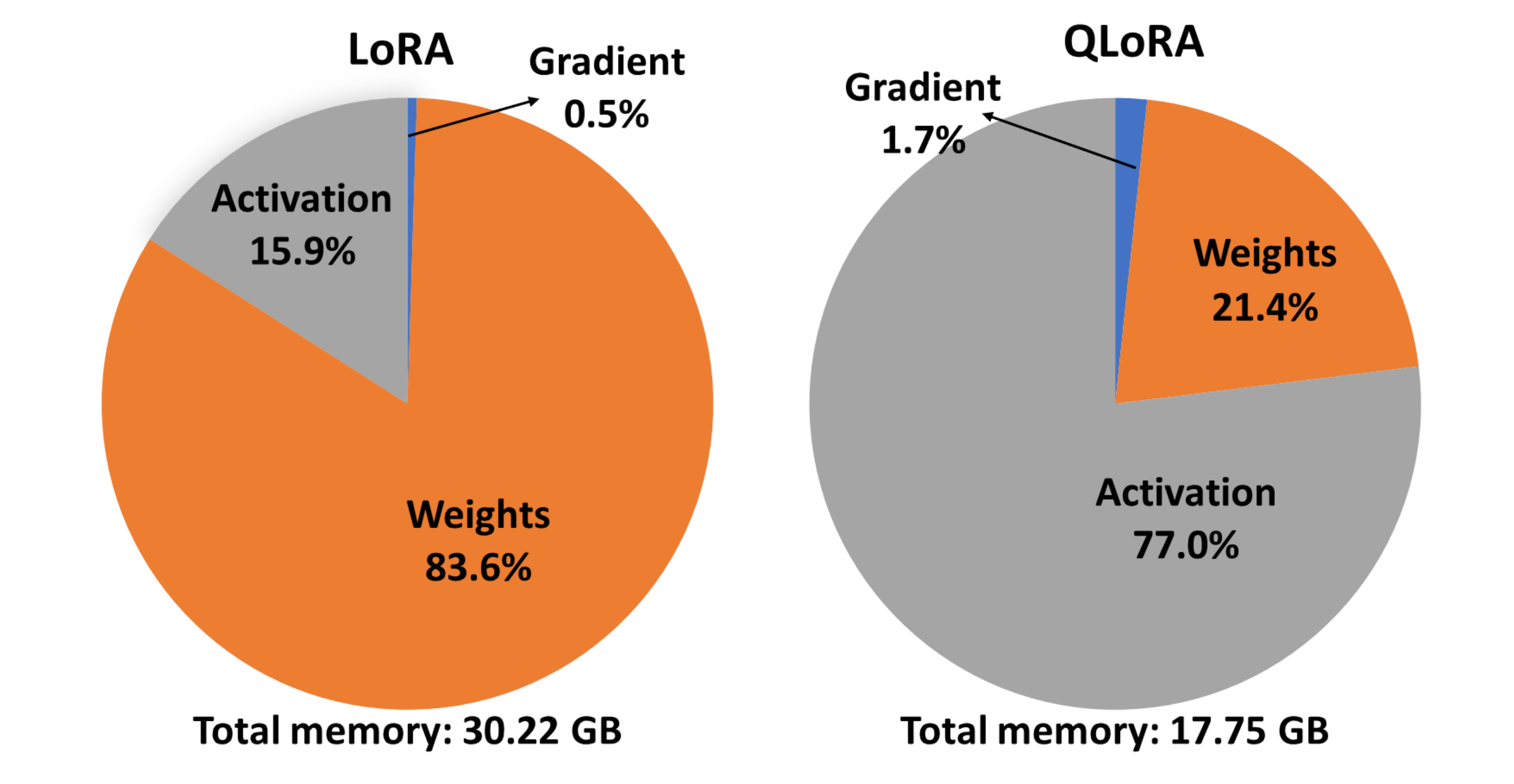}
    \vspace{-1em}
    \caption{Profiling results on the memory footprint when tuning LLaMA-7B with LoRA~\cite{hu2021lora} and QLoRA~\cite{dettmers2023qlora} on the Alpaca~\cite{taori2023stanford} dataset.}
    \vspace{-1.7em}
    \label{fig:profile_mem}
\end{figure}

\textbf{Memory-efficient tuning (MET)} aims to minimize the memory footprint during the tuning process by reducing backpropagation depth, thereby decreasing the number of activations required to be stored in memory~\cite{zhang2023llama,sung2022lst}. Existing MET techniques achieve this goal either using partial tuning to only tune the final few layers~\cite{zhang2023llama} or leveraging side tuning to add a bypass connection between each adapter module with the final output~\cite{sung2022lst}. While the reduction of memory footprint during tuning is highly desirable, existing MET techniques still face an unsatisfactory trade-off between accuracy and memory footprint in LLM tuning. Specifically, for partial tuning, existing attempts on LLMs need to tune more than 80\% of layers of the target LLM to achieve a satisfactory task accuracy~\cite{zhang2023llama}, while side tuning suffers from biased optimization and struggles to achieve task accuracy comparable to SOTA PET techniques~\cite{sung2022lst}. 

\textbf{Compressing-then-tuning} is a series of emerging efficient tuning techniques motivated by the observation that the computation overhead in LLM tuning is dominated by the forward and backward passes of the LLM's backbone, due to the excessive size of the LLM's backbone~\cite{dettmers2023qlora}. Thus, some pioneering works propose to compress the LLM backbone before tuning to \hr{reduce} the computation \hr{and data movement overheads} ~\cite{dettmers2023qlora}. 
However, existing SOTA compressing-then-tuning techniques primarily aim to improve tuning speed, neglecting the extreme memory overhead (e.g., the SOTA compressing-then-tuning method still needs an A100 GPU with 40GB memory to achieve effective tuning on 
\hr{the Llama-70B model~\cite{dettmers2023qlora}).}
This oversight limits the effectiveness of compressing-then-tuning techniques in tuning LLMs on resource-constraint edge devices. 

\vspace{-0.5em}
\subsection{Memory Overhead During Tuning}
\label{sec:profile}
To better understand the gap between the memory needed in existing tuning techniques and the memory available on edge devices, we profile the \textbf{memory} requirements to tune a Llama-7B \hr{model}~\cite{zhang2023llama} with LoRA~\cite{hu2021lora}, one of the SOTA PET techniques, and QLoRA~\cite{dettmers2023qlora}, one of the SOTA compressing-then-tuning techniques, respectively. As shown in Fig.~\ref{fig:profile_mem}, the memory overhead of LoRA is dominated by storing the LLM's backbone weights and the activations for backpropagation. Even after QLoRA compressed the LLM backbone to 4-bit and reduced the overall memory footprint by 41.2\% over LoRA, there remains a \hr{1.48$\times\sim2.22\times$} gap between the memory required for tuning and the memory available on commonly used edge devices (e.g., 8 GB for TX2~\cite{tx2} and 12 GB for Quest Pro~\cite{quest}). 

\vspace{-0.5em}
\subsection{Opportunities for \hr{Efficient} LLM Tuning}
\label{sec:motivation}
\hr{To tackle the aforementioned limitations of existing tuning methods, we identify potential opportunities to improve these methods to develop effective LLM tuning frameworks.}

\hr{On one hand,} to further reduce the \textbf{computation} overhead, we identify a mismatch between the previously successful practice aimed at reducing the model redundancy and the vanilla compression technique used in existing compressing-then-tuning techniques. Specifically, previous efforts (e.g., \cite{yu2022unified} observe that deep learning models exhibit redundancy across different dimensions (e.g., bit-width and sparsity) and at different layers. In contrast, existing compressing-then-tuning techniques often adopt a uniform compression approach, reducing redundancy from only one dimension~\cite{dettmers2023qlora}.

On the other hand, to further reduce the \textbf{memory} overhead, based on our analysis in Sec.~\ref{sec:tuninig}, we summarize that the key to improving the achievable accuracy-memory trade-off lies in the ability to update all layers in the LLM with a limited backpropagation depth. Inspired by the early exit mechanism developed for efficient model inference~\cite{teerapittayanon2016branchynet}, we hypothesize that the outputs from early layers in the LLM can provide meaningful information for prediction. Thus, it is possible to start backpropagation from an early exit layer and still effectively update the model. In this scenario, since backpropagation can be initiated from various early exit layers, the backpropagation depth required for updating all layers in the LLM can be minimized.


\vspace{-0.5em}
\section{Edge-LLM Algorithm}
\subsection{Overview}
\begin{figure}
    \centering
    \includegraphics[width=\linewidth]{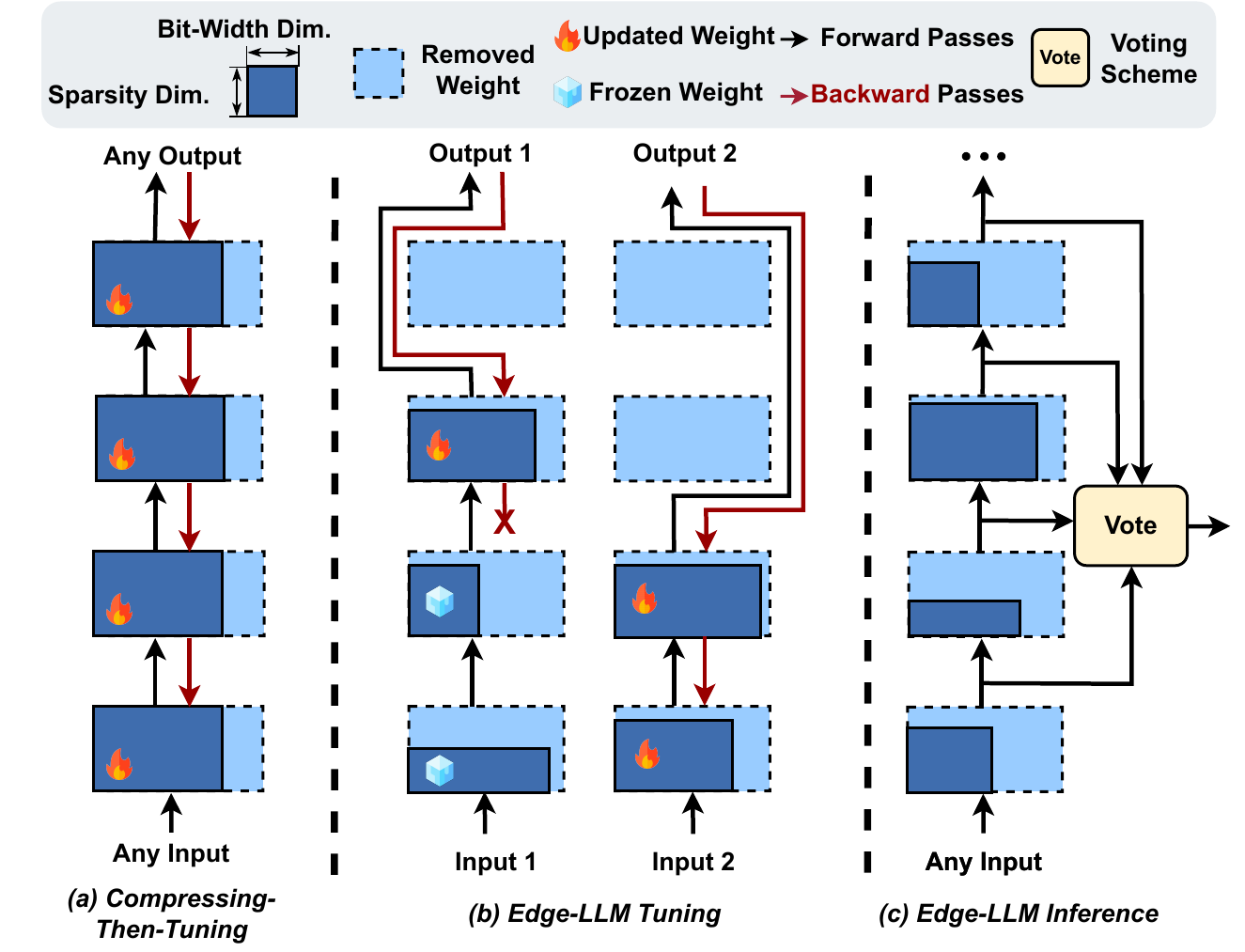}
    \vspace{-2em}
    \caption{\hr{Comparison between (a) the compressing-then-tuning baseline and (b/c) our proposed Edge-LLM method.}}
    \vspace{-1em}
    \label{fig:overview}
\end{figure}

Motivated by the opportunities identified in Sec.~\ref{sec:motivation}, we then introduce the algorithm design of our proposed Edge-LLM framework to facilitate effective 
\hr{and efficient LLM adaptation}
with limited computation and memory overhead. As shown in Fig.~\ref{fig:overview}, our proposed Edge-LLM tuning algorithm integrates two key enablers each leveraging one of the aforementioned opportunities in reducing the computation and memory overhead. Specifically: 
(1) To reduce the \textbf{computation} overhead, we propose \hr{the LUC} technique to diminish the redundancy of the target LLM. This technique is motivated by our empirical observation of 
\hr{the diverse layer-wise sensitivities of LLMs}
to quantization and pruning. Based on the observation above, we develop a low-cost, mean-square-error-based (MSE-based) identifier in LUC to generate a layer-wise compression policy (e.g., layer-wise bit-width and pruning sparsity allocation), aiming to improve the accuracy-efficiency trade-off of LUC over existing compression techniques in compressing-then-tuning frameworks (Sec.~\ref{sec:method_compression}).
(2) To reduce the \textbf{memory} overhead, we propose an adaptive layer tuning scheme that dynamically connects the output of a selected layer (potentially different in each iteration) to the final classification layer with a skip connection during the forward pass. 
\hr{
During backpropagation, only a few preceding layers of the selected layer receive gradient updates. Because the layers selected for updates vary with different inputs, this approach ensures that all layers are effectively updated while minimizing memory overhead. This efficiency is achieved through the reduced depth of backpropagation enabled by the introduction of skip connections.
Furthermore, during inference, we introduce a voting mechanism to enhance the accuracy of LLMs tuned with adaptive layer tuning. This method capitalizes on the ability of adaptively tuned LLMs to produce reasonable outputs from multiple layers. Consequently, each layer generates logits, and a voting process is employed to determine the final output (see Sec.~\ref{sec:method_tuning}).
}

\vspace{-0.5em}
\subsection{Layer-wise Unified Compression (LUC)}
\label{sec:method_compression}
\begin{figure}
    \centering
    \includegraphics[width=0.9\linewidth]{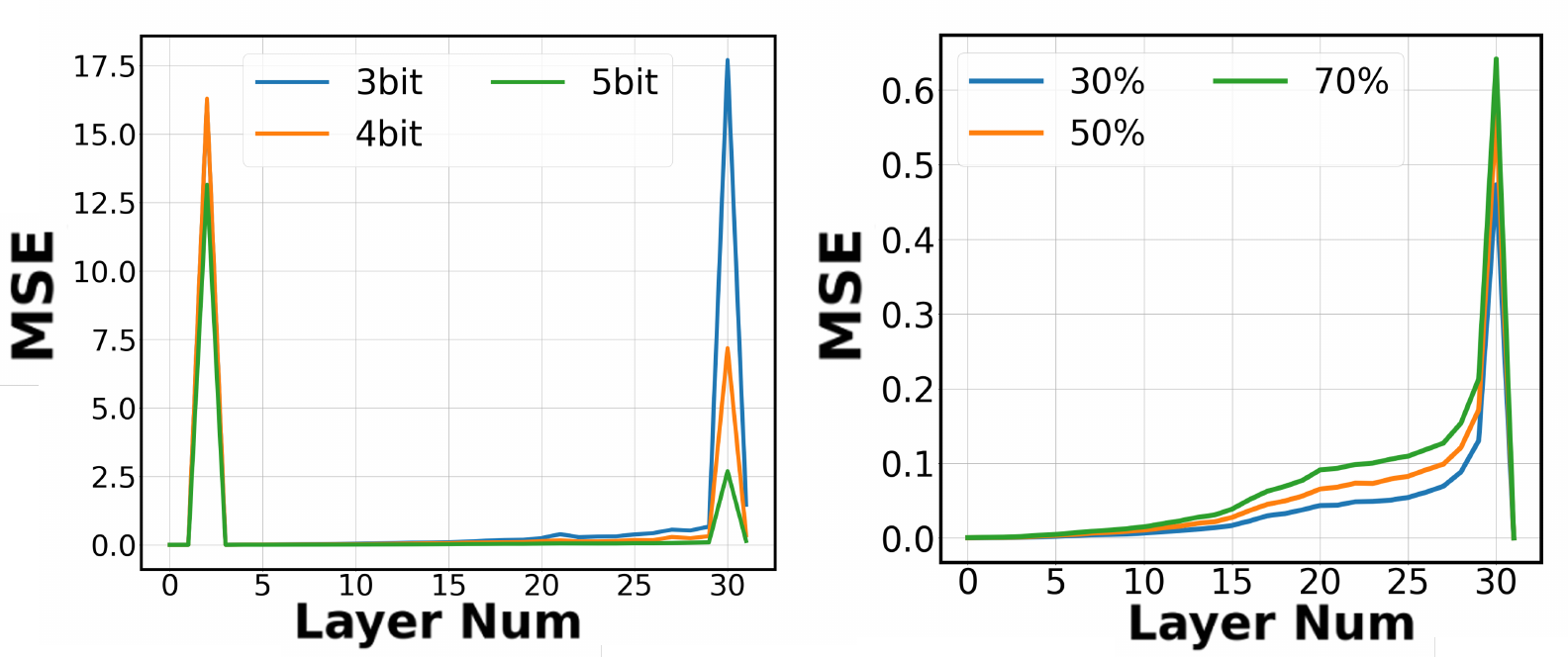}
    \vspace{-1em}
    \caption{Visualization of LLaMA-7B's layer-wise sensitivity to (a) quantization and (b) pruning. }
    \vspace{-1.5em}
    \label{fig:prune_quant_sensitivity}
\end{figure}

\textbf{Motivating observation on LLM's layer-wise sensitivity.} 
In prior studies on model compression, a common understanding is that different layers 
\hr{in a model} 
exhibit different sensitivities to different compression techniques~\cite{yu2022unified}. However, the sensitivities of different layers in LLMs to different compression techniques remain an open question. To address this question, we first explore the layer-wise sensitivities of the target LLM to pruning and quantization. Specifically, we apply different quantization bit-widths and pruning sparsities to each layer of a pretrained 
\hr{LLaMA-7B~\cite{touvron2023llama} model}. By comparing the averaged MSE of the compressed and original layer outputs in the target LLM fed with the same input from the WikiText dataset~\cite{merity2016pointer}, we observe that, as shown in Fig.~\ref{fig:prune_quant_sensitivity}, only a small fraction of layers in the LLM have high sensitivities to compression.


\textbf{Our hypothesis and the proposed LUC.} Based on the observation above, we hypothesize that the high sensitivity (i.e., high MSE) is due to limited redundancy in the corresponding layer, thereby necessitating a lower compression ratio. To this end, we propose the following mapping functions to map the layer-wise MSE to the layer-wise quantization bit-width and pruning sparsity, respectively. For \textbf{quantization}, given an LLM $M$ with $L$ layers, formulating $\mathcal{L}=\{l_0,l_1,\cdots,l_{L-1}\}$, a base quantization bit-width $B$, and the quantization sensitivity (i.e., the MSE between the output of the original layer and the output of the $B$-bit quantized layer) for layer $l_i$ as $s_{quant}^{i}$, we define the optimized quantization bit-width $b_j$ at layer $l_j$ as 
\vspace{-0.6em}
\begin{equation}
    b_j = B + \mathbbm{1}(s_{quant}^{j} \geq \frac{\sum_{i=0}^{L-1}s_{quant}^{i}}{L}),
\vspace{-0.2em}
\end{equation}
where $\mathbbm{1}(.)$ is the indicator function. For \textbf{pruning}, given a target overall pruning sparsity $P$, we define the pruning sparsity $p_j$ at layer $l_j$ as 
\vspace{-1em}
\begin{equation}
    p_j = P\times L\times \frac{s_{prune}^{j}}{\sum_{i=1}^{L-1}s_{prune}^i},
\vspace{-0.3em}
\end{equation}
where $s_{prune}^j$ is the pruning sensitivity for layer $l_j$. 

\vspace{-0.5em}
\subsection{Adaptive Layer Tuning and Voting}
\label{sec:method_tuning}
In this enabler, our objective is to facilitate effective tuning with reduced memory overhead, thereby fitting the tuning process into edge devices with limited memory capacity. To achieve this, the primary challenge we've identified is enabling efficient updates across all layers of the target LLM with restricted backpropagation depth, as analyzed in \hr{Sec.~\ref{sec:motivation}.} 

In Edge-LLM, we alleviate this challenge by constructing a set of exit layers $\mathcal{T} = \{t_0, t_1, \cdots, t_{T-1}\}$. 
\hr{Each exit layer $t_i$ connects to the output of layer $l_{\text{Ceil}((i+1) \times L / T)}$ in the target LLM, functioning as the final output layer.
Note that $T$ represents the number of selectable exit layers, and $L$ denotes the total number of layers in the target LLMs, ensuring that $T<L$.}
In each tuning iteration, we randomly select $t_i \in \mathcal{T}$ as the only exit layer to use, and update the following set of layers $\{l_{\text{Ceil}((i+1) \times L / T)-m},\ l_{\text{Ceil}((i+1) \times L / T)-m+1},\ \cdots, l_{\text{Ceil}((i+1) \times L / T)}, t_i\}$.
\hr{Each layer in this set is equipped with LoRA adapters. Here, $m=\text{Ceil}(L/T)$ denotes the number of layers that have unfrozen trainable parameters in this configuration.}

Furthermore, with the adaptive layer tuning described above, the tuned LLM can generate outputs from all layers $t\in\mathcal{T}$. Although directly using the final output layer $t_{T-1}$ can achieve competitive performance, having multiple available exit layers provides an opportunity to further enhance the performance at inference time by adaptively combining the outputs of different layers. To this end, we propose a voting mechanism to enhance the performance by making predictions based on the outputs from all exit layers. Specifically, inspired by existing findings about the relationship between post-softmax probability and prediction confidence~\cite{pearce2021understanding}, we determine the final output index by choosing the one with the highest post-softmax probability across all exit layers. Specifically, given an output probability matrix $\mathbf{M}$, with each element $\mathbf{m}_{(i,j)}$ representing the output probability for index $j$ from layer  $t_i\in\mathcal{T}$. We first find the location of the maximum value in $\mathbf{M}$ with $(i_{max},j_{max})=\argmax_{i,j}(\mathbf{m}_{(i,j)})$, then we generate the final output as $o=j_{max}$.

\section{Edge-LLM Hardware Scheduling}

\begin{figure*}[t]
    \centering
    \vspace{-1em}
    \includegraphics[width=\linewidth]{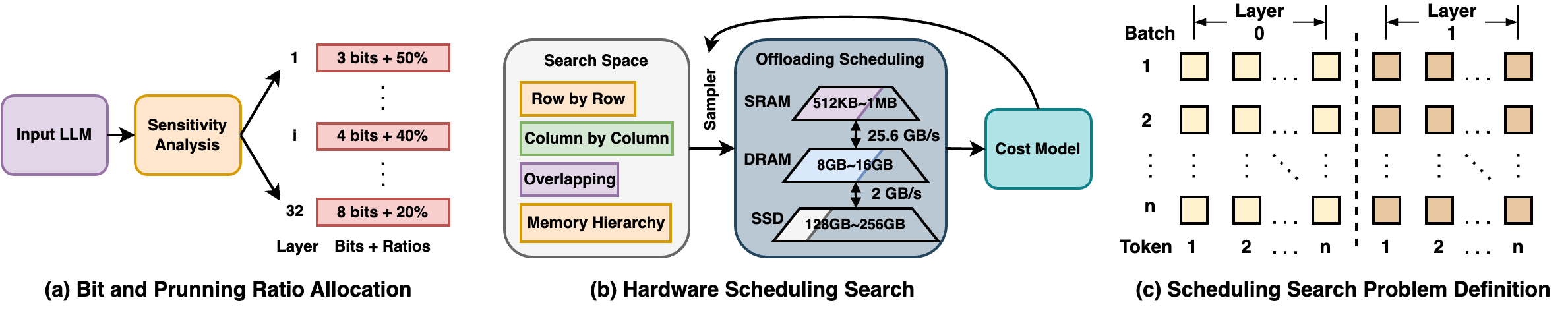}
    \vspace{-2em}
    \caption{The overview of our hardware scheduling. }
    \label{fig:scheduling}
    \vspace{-1em}
\end{figure*}

\textbf{Motivation.}
The aforementioned algorithm designs introduce an irregular computation pattern (i.e., diverse layer-wise quantization bit-width, layer-wise pruning sparsity, and layers to update). This complexity makes it challenging for real devices to fully benefit from the algorithm's theoretical reduction in computation overhead. To address this challenge, we propose a complementary hardware scheduling module, focusing on efficient scheduling and offloading strategies tailored for optimizing LLM inference throughput. The on-chip accelerator SRAM size limitation (512KB$\sim$1MB) highlights the inability to load all model weights and activations, necessitating offloading to secondary storage mediums like DRAM (8GB$\sim$16GB) and SSD (128GB$\sim$256GB). 
\hr{Our hardware acceleration is motivated by the need to establish a comprehensive cost model, serving as the basis for efficient memory scheduling or offloading strategies for each early exit block in the system.}

\vspace{-0.6em}
\subsection{Overview}
\vspace{-0.2em}
In the pursuit of optimizing the scheduling and offloading strategies for LLM hardware accelerators, our methodology allocates bit-widths and pruning sparsities to each layer based on sensitivity (see Sec.~\ref{sec:method_compression}). Subsequently, we conduct a nuanced exploration to identify the optimal offloading strategy for each early exit block.
As depicted in Fig. \ref{fig:scheduling} (a) and (b), these two steps take algorithm hyperparameters as inputs and yield the final allocation strategy and hardware schedulings as outputs.



\vspace{-0.6em}
\subsection{Searching Objective}
\vspace{-0.2em}
We conceptualize the LLM tuning with offloading as a graph traversal problem following \cite{sheng2023flexgen}. In Fig. \ref{fig:scheduling} (c), we present an illustrative computational graph consisting of three dimensions of batches, layers, and tokens. In the depicted graph, each square denotes the computation of a specific layer. Squares sharing the same color indicate the utilization of identical layer weights. A valid path is defined as a trajectory that traverses (i.e., computes) all squares, adhering to the following constraint:

\begin{itemize}
\vspace{-0.3em}
    \item During LLM forwarding or backpropagation, a square's computation depends on the left or right layers in its row being completed, respectively. 

    \item To compute a square, all its inputs (weights, activations, cache) must be loaded onto the on-chip SRAM.

    \item At any given time, the cumulative size of tensors stored on an accelerator must not exceed its memory capacity.
\end{itemize}

The objective is to identify a valid path that minimizes the overall execution time, encompassing both compute costs and I/O costs incurred during the movement of tensors.

\vspace{-0.7em}
\subsection{Block Search Space}
Building upon the aforementioned search objective, we establish a search space encompassing potential valid strategies.

\begin{itemize}
    \item \textbf{Row-by-row.} Existing systems often use solely row-by-row traversal for the activation footprint savings. 
    However, this strategy does not consider the weight sharing between adjacent squares among different bathes, leading to repetitive weight loading I/O costs.

    \item \textbf{Mixed column-by-column and row-by-row.} Alternatively, to reduce I/O costs related to weights, an approach involves traversing the graph column-by-column. This leverages weight sharing among all squares in a column, allowing DRAM preservation for reuse, with activations being loaded and unloaded. As our proposed algorithm techniques can greatly reduce the activation memory footprint requirement, we include mixed column-by-column and row-by-row in search space.
\end{itemize}

\textbf{Considerations.}
\textit{Overlapping.} Another optimization is overlapping. This entails concurrently handling a load of weights for the next layer, the load of activations for the subsequent batch, the storage of activations from the preceding batch, and the computation of the current batch. The integration of overlapping into the block schedule is necessary for delivering the final scheduling.

\textit{Tensor Placement.} In addition to the computation schedule, an effective strategy must delineate the placement of tensors within the memory hierarchy. Three variables, namely $w_{sram}$, $w_{dram}$, and $w_{ssd}$, define the percentages of weights stored on the SRAM, DRAM, and SSD, respectively. Similarly, three variables, $a_{sram}$, $a_{dram}$, and $a_{ssd}$ articulate the percentages of activations; and three variables, $g_{sram}$, $g_{dram}$, and $g_{ssd}$ articulate the percentages of gradients.

\vspace{-0.6em}
\subsection{Cost Models}
Having established the search objective and the search space, the next step is the development of an analytical cost model. 
This model serves the purpose of estimating the execution time based on the specified algorithm parameters and hardware specifications.
The total latency for computing a block can be estimated as $T_{\text{dec}}$.
Assuming perfect overlapping, \(T_{\text{dec}}\) can be estimated as
\begin{equation}
\vspace{-0.2em}
    T_{\text{dec}} = \max(r_{\text{to\_sram}}, w_{\text{to\_dram}}, r_{\text{to\_dram}}, w_{\text{to\_ssd}}, T_{\text{comp}}) 
\vspace{-0.2em}
\end{equation}
where \(r_{\text{to\_sram}}\), \(w_{\text{to\_dram}}\), \(r_{\text{to\_dram}}\), \(w_{\text{to\_ssd}}\), and \(T_{\text{comp}}\) denote the latency of read from DRAM to SRAM, write from SRAM to DRAM, read from SSD to DRAM, write from DRAM to SSD, and computation, respectively, during LLM tuning.

\vspace{-0.5em}
\section{Evaluation}
\subsection{Evaluation Setup}
\textbf{Datasets:} Two commonly used benchmarking dataset including MMLU~\cite{hendrycks2020measuring} and WikiText~\cite{merity2016pointer}. \textbf{Model:} LLaMA-7B~\cite{touvron2023llama}. \textbf{Algorithm baselines:} \hr{The SOTA PET technique, LoRA~\cite{hu2021lora}; the SOTA MET technique, LST~\cite{sung2022lst}; the SOTA compression techniques, Sparse-GPT~\cite{frantar2023sparsegpt} and LLM-QAT~\cite{liu2023llm}; and seven variants of our proposed methods.} \textbf{Hardware baselines:} The SOTA systolic accelerator~\cite{shao2023efficient} dedicated for transformer training.
\textbf{Algorithm implementation:} We use LLM-QAT and Sparse-GPT as the quantization and pruning techniques, respectively, and tune the model following the settings in~\cite{dettmers2023qlora}.
\textbf{Hardware configuration:} The accelerator's DRAM is set to 8GB LPDDR4 and on-chip SRAM to be 1MB, in line with SOTA edge devices~\cite{tx2}, with other hardware configurations following the baseline training accelerator design. \textbf{Evaluation methodology:} We use the SOTA Scale-Sim~\cite{scalesim} simulator to simulate both the baseline accelerator and those after applying our techniques on the baseline accelerator. 

\vspace{-0.5em}
\subsection{Algorithm Evaluation}
To evaluate the performance of our proposed method, we first benchmark our proposed method with existing baseline methods including partial tuning, LST and LoRA tuning on the commonly used MMLU dataset. As shown in Table~\ref{tab:alg_main_exp}, our method consistently achieves a 0.70\%$\sim$1.29\% higher accuracy with the same computation efficiency and a 4$\times$ reduction in memory over the baseline methods. To further validate the key enablers in Edge-LLM, we first evaluate the LUC's perplexity separately on the WikiText-2 dataset over two SOTA compression techniques including SparseGPT and LLM-QAT, and two variants: (1) \underline{Uniform}: using the same quantization bit-width and pruning sparsity across all layers and (2) \underline{Random}: Randomly assign our generated layer-wise pruning sparsities and quantization bits across all layers. As shown in Table~\ref{tab:alg_compress}, our proposed method achieves a 1.28$\sim$2.49 lower perplexity compared to the \underline{Uniform} baseline under similar resource constraints and a 0.50$\sim$1.68 lower perplexity compared to the \underline{Random} baseline under the same efficiency, showing the effectiveness of our proposed LUC. 
\begin{table}[]
    \centering
    \caption{Benchmarking Edge-LLM on \hr{the} MMLU dataset.}
    \vspace{-1em}
    \resizebox{0.8\linewidth}{!}{\begin{tabular}{c|cc|c|c}
    \toprule
        Method & Avg. Bit & Sparsity & Norm. Mem. & MMLU \\
        \midrule
        LoRA & 8.0 & 0\% & 1.00$\times$ & 33.60 \\ 
        \midrule
        Partial Tuning & 5.0 & 50\% & 0.25$\times$ & 30.94\\ 
        Ours & 5.1 & 50\% & 0.25$\times$ &\textbf{31.64}\\ 
        \midrule
        LST & 4.0 & 0\% & 0.29$\times$ & 29.04 \\
        Partial Tuning & 4.0 & 50\% & 0.25$\times$ &28.70\\ 
        Ours & 4.1 & 50\% & 0.25$\times$ &\textbf{29.89} \\ 
        \midrule
        Partial Tuning & 3.0 & 50\% & 0.25$\times$ &26.61 \\ 
        Ours & 3.1 & 50\% & 0.25$\times$ &\textbf{27.68} \\
        \bottomrule
        
    \end{tabular}
    }
    \vspace{-1em}
    \label{tab:alg_main_exp}
\end{table}

\vspace{-0.5em}
\subsection{Hardware Evaluation}
\hr{We evaluate the proposed techniques based on the baseline systolic accelerator designed for transformer training with proper modifications for supporting the proposed techniques~\cite{shao2023efficient}:} 
(1) Since the proposed adaptive layer tuning can be naturally run on the baseline accelerator, there is no need to modify the baseline accelerator;
and (2) For the LUC, we make these modifications: we update the baseline to store the compressed weights on DRAM and SSD. 
\hr{To simplify the design,}
we do not modify the compute core for sparsity and use a simple spatial-temporal flexible-precision MAC unit~\cite{fu20212}. 
We apply our proposed hardware scheduling searching method to find the optimal algorithm-to-hardware mappings. 
Scale-Sim simulation results show that the adaptive layer tuning can achieve 2.24$\times$ speedup; the pruning and adaptive layer tuning can introduce 2.37$\times$ speedup; and combing LUC (4-bit/5-bit) and the adaptive layer tuning can give 3.38$\times$/2.92$\times$ overall speedup, respectively. 

\vspace{-0.5em}
\section{Conclusion}
In this paper, we introduce an LLM tuning framework, Edge-LLM, achieving efficient LLM adaptation on edge devices. Experiments demonstrate that Edge-LLM achieves efficient adaptation with comparable performance as vanilla tuning with a 2.92$\times$ speed up and a 4$\times$ memory reduction.

\begin{table}[]
    \centering
    \caption{Ablation on LUC's performance with its variants}
    \vspace{-0.8em}
    \resizebox{0.65\linewidth}{!}{\begin{tabular}{c|cc|c}
    \toprule
        Method & Avg. Bit & Sparsity & Perplexity \\
        \midrule
        SparseGPT & 8.0 & 50\% &15.88\\ 
        LLM-QAT & 8.0 & 0\% &13.34\\
        \midrule
        Uniform & 5.0 & 50\%  & 17.61 \\ 
        Random & 5.1 & 50\%  & 16.21 \\ 
        Ours & 5.1 & 50\%  & \textbf{15.71}\\ 
        \midrule
        Uniform & 4.0 & 50\%  & 19.86 \\ 
        Random & 4.1 & 50\%  & 19.81 \\ 
        Ours & 4.1 & 50\% & \textbf{18.58}\\ 
        \midrule
        Uniform & 3.0 & 50\%  & 32.52\\ 
        Random & 3.1 & 50\%  & 31.71 \\ 
        Ours & 3.1 & 50\%  &\textbf{30.03}\\
        \bottomrule
    \end{tabular}
    }
    \vspace{-1.2em}
    \label{tab:alg_compress}
\end{table}
\vspace{-0.5em}
\section*{Acknowledgement}
This work was supported in part by CoCoSys, one of the seven centers in JUMP 2.0, a Semiconductor Research Corporation (SRC) program sponsored by DARPA, and the National Science Foundation (NSF) through the NSF CAREER funding (Award number: 2048183).

\vspace{-0.5em}
\bibliographystyle{ACM-Reference-Format}
\bibliography{sample-base}

\end{document}